\documentclass[letterpaper, 10 pt, conference]{ieeeconf}

\IEEEoverridecommandlockouts                             
\usepackage{graphicx}       
\usepackage{dblfloatfix}    
\usepackage{hyperref}       
\usepackage{orcidlink}      
\usepackage[all]{nowidow}   

\let\labelindent\relax
\usepackage{enumitem}       
\setitemize{leftmargin=*}

\usepackage{array}          
\newcolumntype{L}[1]{>{\raggedright\let\newline\\\arraybackslash
\hspace{0pt}}m{#1}}        

\newcommand\copyrighttext{%
    \footnotesize \copyright{ }2024 IEEE. Personal use of this material is permitted. Permission from IEEE must be obtained for all other uses, in any current or future media, including reprinting/republishing this material for advertising or promotional purposes, creating new collective works, for resale or redistribution to servers or lists, or reuse of any copyrighted component of this work in other works.}
\newcommand\copyrightnotice{%
    \begin{tikzpicture}[remember picture,overlay]
    \node[anchor=south,yshift=15pt,xshift=0pt] at (current page.south) {\parbox{\dimexpr\textwidth-\fboxsep-\fboxrule\relax}{\copyrighttext}};
    \end{tikzpicture}%
}

\title{\LARGE \bf
CARLOS: An Open, Modular, and Scalable Simulation Framework for the Development and Testing of Software for C-ITS
}

\author{Christian Geller\orcidlink{0000-0001-8655-3201}, Benedikt Haas\orcidlink{0009-0004-4842-9997}, Amarin Kloeker\orcidlink{0000-0003-4984-2797}, Jona Hermens\orcidlink{0009-0005-9193-3336}, \\ Bastian Lampe\orcidlink{0000-0002-4414-6947}, Till Beemelmanns\orcidlink{0000-0002-2129-4082}, Lutz Eckstein
\thanks{*This research is accomplished within the project ”AUTOtech.\textit{agil}” (FKZ 01IS22088A). We acknowledge the financial support for the project by the Federal Ministry of Education and Research of Germany (BMBF).}
\thanks{All authors are with the research area Vehicle Intelligence \& Automated Driving, Institute for Automotive Engineering, RWTH Aachen University, Germany 
\tt\small\{firstname.lastname\}@ika.rwth-aachen.de
}
\vspace{-0.4cm}
}

\begin{document}

\bstctlcite{IEEEexample:BSTcontrol}

\maketitle

\thispagestyle{empty}
\pagestyle{empty}

\copyrightnotice

\begin{abstract} 

Future mobility systems and their components are increasingly defined by their software. The complexity of these cooperative intelligent transport systems (C-ITS)  and the ever-changing requirements posed at the software require continual software updates. The dynamic nature of the system and the practically innumerable scenarios in which different software components work together necessitate efficient and automated development and testing procedures that use simulations as one core methodology. The availability of such simulation architectures is a common interest among many stakeholders, especially in the field of automated driving. That is why we propose CARLOS - an open, modular, and scalable simulation framework for the development and testing of software in \mbox{C-ITS} that leverages the rich CARLA and ROS ecosystems. We provide core building blocks for this framework and explain how it can be used and extended by the community. Its architecture builds upon modern microservice and DevOps principles such as containerization and continuous integration. In our paper, we motivate the architecture by describing important design principles and showcasing three major use cases - software prototyping, data-driven development, and automated testing. We make CARLOS and example implementations of the three use cases publicly available at \href{https://github.com/ika-rwth-aachen/carlos}{github.com/ika-rwth-aachen/carlos}.

\end{abstract}

\section{INTRODUCTION} 
\label{sec:introduction}


Simulations play a vital role in the development and testing of automated driving functions~\cite{Groh19}. Future cooperative intelligent transport systems (C-ITS) will increase their necessity even further due to the large number of involved entities and their complex interactions. Whether requirements posed at the security and safety of software are met must be tested before software is deployed. Simulations help deal with this task by being able to run vast amounts of relevant scenarios in which different software components may interact. Although simulations are affected by a reality gap, continuous advancements in their fidelity enable them to play a more central role in software development and testing~\cite{Kloeker23}\cite{Huch23}\cite{Zhong21}. Thus, simulative testing represents an expanding topic in both research and development, as seen in research projects such as \mbox{AUTOtech.\textit{agil}}~\cite{VanKempen23}.

Numerous stakeholders have therefore developed solutions for simulation frameworks and architectures~\cite{Stepanyants23}. While progress has been made, current solutions often lack in scalability, usability, maintainability, and other key aspects. Most importantly, existing solutions are often tailored to the specific needs of their developers, may not be open-source, or not modular enough to be easily extended by the community.

\begin{figure}[t]
    \centering
    \vspace{1.8mm}
    \includegraphics[width=\linewidth]{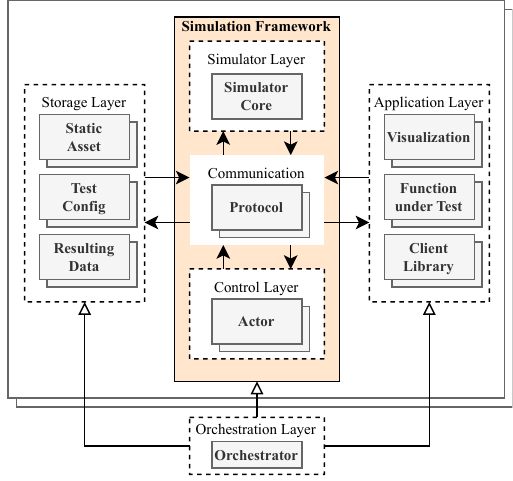}
    \caption{Overview of the proposed modular simulation architecture. It divides functionality into multiple layers, resulting in a generic microservice architecture that is flexible for multiple use cases. A managing orchestration layer enables automation and scalability of simulations.}
    \label{fig:introduction:architecture}
    \vspace{-1mm}
\end{figure}

Thus, an open, modular, and scalable simulation architecture is beneficial for the development and testing of automated driving software components. In this paper, we propose a generic architecture that is abstractly depicted in Fig.~\ref{fig:introduction:architecture}. Aligned with recent trends toward containerization and microservices, this novel architecture is characterized by modularity and flexibility. In detail, a separation into functional components enables dynamic adaptation to various simulation use cases and a seamless integration with DevOps workflows such as continuous integration (CI). Moreover, common orchestration tools can achieve the scalability of simulations to multiple concurrent runs, which significantly enhances testing throughput and efficiency.

As a reference implementation, we present CARLOS, a novel open-source framework incorporating the proposed simulation architecture within the CARLA ecosystem~\cite{Dosovitskiy17}. CARLOS and comprehensive demonstrations for different use cases are made publicly accessible on GitHub. Consequently, we provide a strong foundation for further development driven by the CARLA and ROS community.

\vspace{0.05cm}
\noindent The main contributions can be summarized as follows:

\begin{itemize}
    \item Use case analysis for simulative testing in the context of automated driving.
    \item Conceptualization of a modular simulation architecture using common standards.
    \item Implementation of CARLOS, a containerized framework using the open-source CARLA ecosystem.
    \item Illustrative provision of examples for specific use cases, allowing simple adaptation.
\end{itemize}


\section{STATE OF THE ART} 
\label{sec:sota}

\subsection{Microservice Architectures} 
\label{sec:sota:microservices}

Microservice architectures can be understood as a successor of the service-oriented architecture (SOA), a well-established architectural style that is often seen as the opposite and a competitor of the monolithic architecture. This is due to it solving many of the limitations and shortcomings of the monolith approach, some of those being maintainability, scalability, and reusability~\cite{Amazon24}. More recently, the microservice architecture has gained in popularity, as it shares many of SOA's strengths by following a similar architectural approach while also deviating at a few key points. For instance, it furthers independence and autonomy of services by splitting up the data store and allowing each service to be coupled with its own, contrasting the SOA approach where storage is shared between services. Additionally, instead of relying on a complex, shared service bus for communication, which is often seen in SOAs, microservices instead employ service-level APIs and gateways for external and inter-service communication~\cite{Jamshidi18}.

There are, of course, some major challenges that come with microservices. To facilitate the flexibility and reproducibility of deployments across a multitude of platforms, virtualization and containerization have emerged, the latter of which being more lightweight, and thus more fitting for microservices~\cite{Tasci18}.
There is also the issue of increased complexity in orchestration and management, as well as the need for extensive monitoring of such often distributed systems~\cite{Baskarada20}.
This is where Docker and Kubernetes respectively have established themselves as the de-facto industry standard~\cite{Hardikar21}, while also having notable and valuable alternatives like Podman or Docker Swarm.

The traits and technologies of microservices are highly desirable for many fields, especially in the booming space of cloud computing. The increasing shift towards cloud computing, and thus towards microservices and SOA, is also felt in robotics~\cite{Lampe23}\cite{Xia18}\cite{Tasci18}, where especially the C-ITS sector shows interest. Besides potentially shifting to the cloud and utilizing containerization and orchestration for live systems, these tools have also shown to be exceptionally useful for the development and testing of C-ITS software~\cite{White17}\cite{Busch23}\cite{RedHat24}. 


\subsection{Test Design} 
\label{sec:sota:testing}

Testing has historically been deeply connected to the process of developing software due to the close relationship between systems engineering and modern software engineering~\cite{Forsberg92}. Besides more recent trends in the software development life cycle (SDLC) like agile development, the V model is part of the more fundamental results of systems engineering~\cite{Akinsola20}. It introduces different levels of testing that changed and expanded over the years but classically breaks down to unit, integration, system, and acceptance testing.

These levels can also be found in SOA and microservices, where the high modularity due to a separation into smaller services leads to an increased need for and interest in integration and unit testing, as well as automation of tests~\cite{Waseem20}. This is especially true for automated driving systems (ADS), in which highly critical software components need to be thoroughly tested to guarantee safety. In practice, this has proven to be exceptionally difficult, not only due to the high risk and low reproducibility of tests on public roads. In detail, an excessive amount of random testing kilometers would be statistically required to guarantee a human-comparable safety level~\cite{Wachenfeld2016}. To address these challenges, several other approaches for ADS testing have emerged, with some of the more promising being simulation and scenario-based testing~\cite{Lou22}\cite{Fremont20}.

The latter abstracts actual or potential traffic into formalized scenario descriptions that can be tested systematically. Expert scenario approaches are handcrafted or systematically derived~\cite{Weber23}. Additionally, scenario databases can be accumulated from real-world driving datasets~\cite{Li23}. The {scenario.center}~\cite{Ika24} combines both approaches and uses real trajectory data to analyze the occurrence and concrete expressions of a comprehensive set of scenarios. Subsequently, concrete scenarios are stored in scenario databases using standards such as OpenSCENARIO~\cite{Asam24} or Scenic~\cite{Fremont18}.

A variety of test benches can be used to conduct scenario-based tests, including simulations, hardware-in-the-loop, proving ground, or field operational tests~\cite{Steimle22}. Simulative testing offers a cost-effective and thus efficient method as it replicates the environment, drivers, vehicles, or sensors in a detailed manner~\cite{Fremont20}. This ensures that tests are conducted in a controlled and safe environment without creating an actual hazard. In addition, simulation tests are renowned for their reproducibility and ability to perform a huge amount of tests in parallel~\cite{Dona22}.

\subsection{Simulation Frameworks} 
\label{sec:sota:simulation}

Various simulation frameworks exist for conducting simulated tests, each focusing on certain aspects and different levels of detail~\cite{Stepanyants23}. A widely used simulation framework is the CARLA simulator~\cite{Dosovitskiy17}, which features highly accurate graphics, realistic sensor modeling, and multiple standardized interfaces. Additionally, CARLA is open-source, actively developed, and supported by a large community. The CARLA ecosystem includes multiple additional components for executing scenarios or bridging to the widely-used Robot Operating System (ROS)~\cite{Macenski22}. It also supports harmonized standards, e.g., from the ASAM OpenX ecosystem~\cite{Asam24}. Other simulation frameworks tending to high simulation fidelity are Virtual Test Drive and CarMaker, the latter with a focus on vehicle dynamics. In addition, a powerful testing tool suite, VIRTO, exists to enable continuous simulative testing with CarMaker on a large scale in cloud environments. On the other side, frameworks like OpenPASS~\cite{Dobberstein2017}, esmini, or CommonRoad~\cite{Althoff17} focus on lightweight scenario execution in line with harmonized interfaces. This mainly increases the applicability for testing entire scenario databases in a continuous safety assurance process.

Many of the simulation frameworks mentioned above are limited in their capabilities or not open-source. In addition, most of them primarily offer a fundamental simulation core but not necessarily an entire architecture to perform effective simulations during development and testing. However, there are extended and more advanced frameworks enabling comprehensive testing processes. As an example, automatic adversarial testing approaches aim to identify potential failures systematically within simulations~\cite{Ramakrishna22}\cite{Tuncali18}. In order to cover a variety of simulation use cases apart from testing, a flexible and modular architecture is crucial, consisting of several powerful, interacting components~\cite{Saigol18}. However, the existing modular simulation architectures cover only specific use cases, but can not flexibly be expanded or scaled, which imposes the necessity to conceptualize a novel, more generic simulation architecture for the continuous development and testing process in automated driving.

\section{ANALYSIS} 
\label{sec:analysis}

The following examples of various simulation use cases motivate the necessity of a novel and generic simulation architecture. In addition, we derive explicit core design principles for the envisaged architecture to overcome existing shortcomings within other architectures.

\subsection{Use Case Analysis}
\label{sec:analysis:use-cases}

\begin{figure}[b]
    \centering
    \includegraphics[width=1\linewidth]{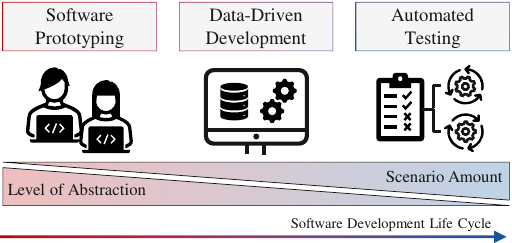}
    \caption{Exemplaric use cases within the software development process motivating the usage of a generic and modular simulation architecture. The scenario amount and level of abstraction vary across the different use cases.}
    \label{fig:analysis:use-cases}
\end{figure}

In automated driving, simulations have a wide range of use cases, aiming at initial developing phases but also testing procedures. Hence, we describe and analyze three fundamental use cases, starting with initial software prototyping, a comprehensive data-driven development, and subsequent automated testing processes, visualized in Fig.~\ref{fig:analysis:use-cases}.

\renewcommand{\arraystretch}{0.5}
\begin{figure*}[b]
    \fontsize{9pt}{9pt}\selectfont
    \begin{center}
        \begin{tabular}{L{4cm} L{4cm} L{4cm} L{4cm}}
        \textbf{Simulation Core} & \textbf{Standardized Interfaces} & \textbf{Modular Containerization} & \textbf{Automated Testing} \\ \\
        \hline
        \begin{itemize}
            \item high simulation fidelity minimizing the reality gap
            \item comprehensive data availability
            \item performance up to real-time capability
        \end{itemize} &
        
        \begin{itemize}
            \item customizable and open data interfaces
            \item Robot Operating System communication interfaces
            \item ASAM OpenX scenario interpretation
        \end{itemize} & 
        
        \begin{itemize}
            \item modularization following separation of concerns
            \item minimal and powerful Docker images
            \item maintainability and updatability
        \end{itemize} & 
        
        \begin{itemize}
            \item automation within CI pipelines
            \item scalability of multiple simulations
            \item automatic scenario evaluations
        \end{itemize} \\
        
        \end{tabular}
    \end{center}
    \caption{Derived requirements for a generic simulation architecture covering the selected use cases in development and testing of automated vehicles.}
    \label{fig:analysis:requirements}
\end{figure*}

\subsubsection{Software Prototyping}

The first use case addresses the initial development phase of an automated driving function, where developers aim for simple and quick prototyping. It focuses on the flawless integrability of the component under test and ensures general operability within the overall system. An early integration facilitates interface harmonization, which is particularly important when assembling modules from different entities. In many cases, a single example scenario is sufficient to carry out initial tests. However, high simulation fidelity is required in many cases, which implies realistic simulation conditions, ranging from realistic environments to accurate sensor, vehicle, or driver models.

Some simple examples in initial development phases are open-loop tests of perception modules checking general functionality or sensor placement experiments to produce a variety of possible sensor setups. Supplementary, also closed-loop tests can be performed with more advanced software stacks to verify interface implementations. Using multiple vehicle instances enables the development of cooperative functionalities. Thus, a variety of simulation possibilities exist even in early development, many of which demand comprehensive input data and harmonized interfaces. In addition, a simple integration of and flexible interaction with novel functionalities is crucial.

\subsubsection{Data-Driven Development}

The second use case covers development processes using large amounts of data, which are not effectively obtainable in real-world settings, thus motivating simulations. For many applications, simulative data can be sufficiently accurate to be integrated into the data-driven development process~\cite{Bewley19}. This includes training data for machine learning algorithms but also closed-loop reinforcement learning. Potentially interesting data includes raw sensor but also dynamic vehicle data in a wide variety at large scale. Simulations additionally enable data generation beyond the physical limits of vehicle dynamics or sensor configurations. To accumulate large amounts of data, relevant simulation parameters can be automatically sampled along different dimensions. Subsequently, automation and parallelization empower a cost-effective execution of multiple simulations, especially when using already established orchestration tools.

\subsubsection{Automated Testing}

In the third use case, simulation is considered to systematically evaluate a large number of defined tests, potentially within the safety assurance process. A specific test configuration may encompass both a concrete scenario and well-defined test metrics for evaluation. Thus, a direct interface to a standardized scenario database is favorable, and custom pass-fail criteria need to be configurable to deduce objective test results. Scalability drastically improves efficiency when simulating multiple test configurations. Moreover, embedding the simulation architecture in a CI process further accelerates the entire safety assurance.

Typically, only a few rough scenarios are considered during initial prototyping, but the scenario amount and level of detail increase when it comes to automated test processes. This also refers to the \textit{level of abstraction} established for scenarios within the PEGASUS~\cite{DLR24} project and can be adapted to the use cases in a similar manner. Fig.~\ref{fig:analysis:use-cases} demonstrates the variety in scenario amount and level of detail across the various use cases. Additionally, the use cases can be assigned to different temporal stages within the SDLC.

\subsection{Requirement Analysis} 
\label{sec:analysis:requirements}

Following the described use cases, specific requirements are derived, categorized, and incorporated into the conceptual design of the simulation architecture as well as for subsequent evaluations. Fig.~\ref{fig:analysis:requirements} gives a brief overview of the gathered requirements.

Essential conditions are imposed on the central \textit{simulation core} since it needs to model reality with optimal performance and high accuracy. This implies accurate data generation and direct provision in every simulation timestep. The usage of \textit{standardized interfaces} can be simplified using the ROS ecosystem for data processing. In addition, compatibility with harmonized ASAM OpenX interfaces is a fundamental requirement, especially when evaluating scenario-based tests in simulations. Beyond that, a flexible and \textit{modular containerization} is a key pillar of a generic simulation architecture and is in line with the trends towards SOAs and microservices, increasing both maintainability and the updateability of the system. Furthermore, containerization allows for orchestration, which not only improves \textit{automated testing} but also enables efficient scaling to concurrent simulations, possibly even distributed across different hosts. Further, the implementation of automated evaluation procedures accelerates the overall simulation process of automated driving functions at large scale.

\section{ARCHITECTURE} 
\label{sec:architecture}

The analysis in Sec.~\ref{sec:analysis} showed that the increased focus on particular types of use cases necessarily leads to specific requirements imposed upon the simulation architecture, which can be investigated in Fig.~\ref{fig:analysis:requirements}. Features like modularity, scalability, and customizability are notably absent in many traditional C-ITS simulation frameworks~\cite{Kirchhof19}. This necessitates an architecture that enriches a simulator core with said features to properly cover the focused use cases. In this section, we propose and describe such an architecture that meets the aforementioned requirements and is widely applicable due to its level of abstraction and, thus, high flexibility. 

Our architecture consists of several \textit{layers} that encapsulate different components with aligned purposes. An overview of that architecture is shown in Fig.~\ref{fig:introduction:architecture}. 

The main layers are the following:
\begin{itemize}
    \item \textbf{Simulation layer} - simulator combined with additional interfaces and capabilities;
    \item \textbf{Storage layer} - persistent data used for or generated by the simulation layer;
    \item \textbf{Application layer} - software/users interacting with the simulation layer to achieve certain goals.
\end{itemize}

\noindent Furthermore, the simulation is composed of \textit{sub-layers}, namely the simulator layer housing the core that performs the simulation work, the interface layer containing protocols for interaction with the simulator, and the control layer which includes a set of actors that extend the functionalities of the core. Letting each instance of the simulation have its own set of components inside of these \textit{sub-layers} improves customizability and flexibility, as it allows users to tailor it to their concrete use cases.
In following a microservice approach by separating the simulation layer into different components, we increase modularity and thus gain the earlier discussed benefits like better maintainability. 
This is especially true if said components are containerized, which opens up the architecture to common orchestration tools and vastly simplifies a distributed simulation. The architecture captures this by including an \textit{orchestration layer} with an orchestrator that dynamically manages all other layers.

\subsection{Components} 

After establishing this abstract architecture, we now introduce CARLOS, a framework that implements this architecture to enhance CARLA as a simulator core and that has proven itself during extensive usage in our research and work. We chose CARLA as the core of our simulation framework due to the benefits described in Sec.~\ref{sec:sota}, like its high maturity and simulation fidelity, as well as its rich ecosystem with additional components. Some of these components are crucial to our framework, thus a brief overview of them and our additional changes follows. All code modifications and prebuilt Docker images of the components are made publicly available on GitHub and Docker Hub.

\textbf{1) Simulation Core:} The \textbf{CARLA server}
constitutes the central element of the framework and handles all graphical and dynamic calculations in the individual simulation time steps. Within our GitHub repository, we extend the pre-existing Dockerfiles to create enhanced Ubuntu-based container images of CARLA via novel CI pipelines.

\textbf{2) Communication Actor:} The \textbf{ROS bridge}
is the component that facilitates the powerful combination of CARLA and ROS. It retrieves data from the simulation to publish it on ROS topics while simultaneously listening on different topics for requested actions, which are translated to commands to be executed in CARLA. This is realized by utilizing both RPC via the CARLA Python API, as well as the ROS middleware, in tandem. 

Since CARLOS focuses on \mbox{ROS 2} (from now on also referred to as ROS), DDS is the middleware used here. Additionally, we provide modernized container images through our CI, where texttt{docker-ros}~\cite{Busch23} enables a continual building of said images with recent versions of ROS, Python, and Ubuntu.

\textbf{2) Control Actor:}
To enable scenario-based testing and evaluation, the \textbf{Scenario Runner} is used. It is a powerful engine that follows the OpenSCENARIO standard for scenario definitions. An additional ROS service allows other ROS nodes to dynamically issue the execution of a specified scenario using the mentioned Scenario Runner. For the creation of more modern and lightweight container images, a custom Dockerfile is published alongside this paper.

\subsection{Data Generation} 

In Sec.~\ref{sec:analysis}, we described data generation for data-driven development as one fundamental use case of simulations. While most simulators offer extensive configuration options, they often lack a methodology for systematically conducting a series of simulations with varying configurations to generate large datasets. This is particularly necessary for generating perception or driving data through simulation. Consequently, we have developed a data generation pipeline for CARLOS to address this need. The objective is to efficiently and easily simulate all selected permutations of parameters across a wide parameter space and store the resulting data. The framework is designed to be as flexible and scalable as possible in executing these simulations.

Our pipeline code is written in Python and uses ROS for the communication with the simulation core. This enables straightforward modifications of the pipeline and its adaptation to other simulation cores. Hence, configuring the pipeline requires only a basic JSON file. The settings in this file are divided into two parts: general simulation settings and the parameter space for generating all permutations with which the simulations are conducted.

Within the general settings, for instance, among numerous other options, it is possible to specify the length of a simulation run, to determine which services to initiate, and to select which ROS topics to capture. The pipeline also allows the execution of simulation-independent services if needed, such as converting ROS data into other formats.

The second part of the configuration file can encompass the parameter space from which all permutations are generated. This space is freely expandable and is only limited by the capabilities of the simulation core. Parameters might include sensor positioning, maps, scenarios, and more. Additionally, it is also possible to provide concrete \mbox{OpenSCENARIO} definitions as a list of scenario files.

A complete list of configuration options and the implementation itself can be found in the CARLOS GitHub repository.

\section{EVALUATION} 
\label{sec:evaluation}

Our simulation architecture is designed based on fundamental requirements derived in Sec.~\ref{sec:analysis}. These requirements are now incorporated to evaluate the architecture in a structured manner. Subsequently, we evaluate a specific implementation of the architecture in an exemplary use case.

\subsection{Architecture Evaluation}
\label{sec:evaluation:architecture-evaluation}

\begin{figure}[b]
    \centering
    \includegraphics[scale=1]{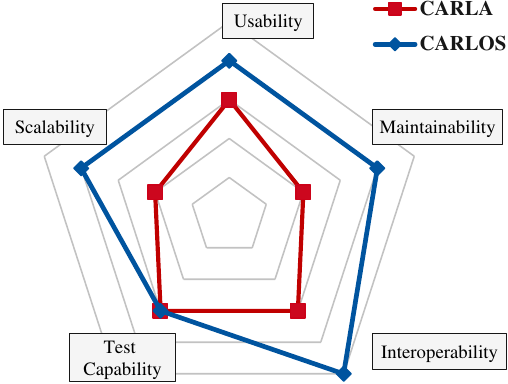}
    \caption{Qualitative evaluation of multiple capabilities by comparing our proposed simulation framework CARLOS with the native CARLA ecosystem. The detailed assessment is based on pre-defined metrics and in-house user assessments.}
    \label{fig:evaluation:architecture-comparison}
\end{figure}

A number of derived requirements, in particular the simulation realism or real-time capability, depend on the simulation core itself. However, our general architectural design allows the core to be flexible and exchangeable, which is why the architecture itself is evaluated using only simulation core-independent evaluation metrics. Thus, we evaluate our open-source simulation framework CARLOS against essential capabilities and provide a qualitative comparison against the native CARLA ecosystem. 

$\bullet$ \textbf{Usability} describes the simplicity of interacting with a simulation architecture to achieve goals effectively and satisfactorily. Well-documented instructions, meaningful examples, and a large community generally enhance usability. All enrichments provided within the CARLA ecosystem remain available within our framework. Additionally, we offer well-documented demonstrations for all described use cases, assisting in the first steps when using the containerized framework and generally improving usability.

$\bullet$ \textbf{Maintainability} refers to the ability to customize the architecture and exchange or update individual components independently. Thus, a modular and containerized structure is essential. Besides, open-source software bypasses potential shortcomings posed by license regulations. Thanks to modular containerization, our open framework offers an ideal basis for continual updates of an individual module. Additionally, all Docker images are built within GitHub CI pipelines, which enables increased maintainability compared to the native CARLA system. 

$\bullet$ \textbf{Interoperability} defines the capability to bind custom functions to the existing architecture. This includes support for standardized and custom interfaces but also a containerized integration. CARLA already offers a number of valuable interfaces, such as ROS or OpenX, which simplify the integration of custom functions. Nevertheless, the microservice architecture of CARLOS facilitates the integration of software functions even more by using existing container orchestration tools like Docker Compose and enabling the use of more sophisticated tools like Kubernetes. Thus, complex software systems composed of many different containers can be easily integrated into our proposed simulation framework.

$\bullet$ \textbf{Scalability} and automation are further key challenges during development and testing. Compared to the native CARLA ecosystem and accelerated by the nature of a SOA, CARLOS focuses on the automation of simulations using common container orchestration tools and CI pipeline integrations. As an additional example, multiple simulations are conducted sequentially within the data generation pipeline. Parallel and distributed simulations become feasible when using powerful orchestration tools. Thus, the novel framework CARLOS significantly increases scalability.

$\bullet$ \textbf{Test Capability} is a decisive factor in the automatic evaluation process of simulations within safety assurance. All powerful existing CARLA evaluation methods and metrics can still be used in our containerized framework design. In addition, custom evaluation modules can be integrated as novel components, depending on the corresponding simulation use case.

Fig.~\ref{fig:evaluation:architecture-comparison} visualizes the comparison of our new proposed architecture and the baseline CARLA ecosystem. It shows that CARLOS can substantially enhance maintainability, interoperability, and scalability, which is crucial in agile development and continuous testing processes.


\subsection{Use Case Evaluation}
\label{sec:evaluation:use-case-evaluation}

We can demonstrate the core capabilities of CARLOS most effectively based on an exemplary use case. Specifically, we aim to evaluate a novel, containerized C-ITS perception module within a concrete simulation scenario. In detail, the module involves a sensor data processing algorithm that takes lidar point clouds as input to detect 3D objects. Thus, we aim to integrate such a point cloud object detection component within the simulation framework and evaluate the component's performance by comparing the resulting object data with the simulated objects. For simplicity, the component is implemented within the ROS ecosystem, enabling a straightforward integration via harmonized data interfaces.

\begin{figure}[t]
    \centering
    \vspace{1.8mm}
    \includegraphics[scale=1]{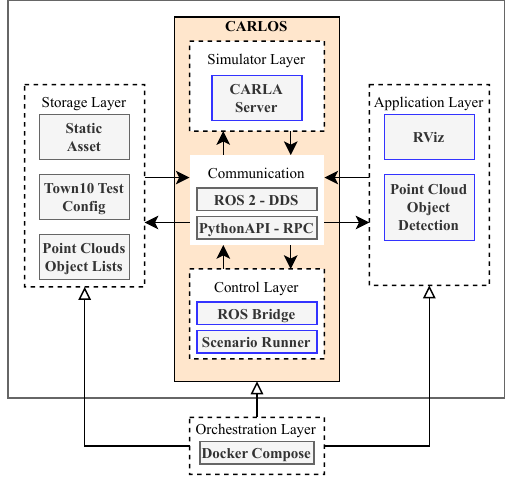}
    \caption{Specialized simulation framework CARLOS, focusing on the integration of a point cloud object detection module. Docker containers are indicated by blue frames.}
    \label{fig:evaluation:use-case-architecture}
\end{figure}

For this specific use case, the framework comprises a {CARLA server}, a {ROS bridge} facilitating the transfer into the ROS ecosystem, and a {Scenario Runner} to control the scenario. Visualization and user interaction are achieved through the ROS tool {RViz}. All mentioned simulation components, including the novel {point cloud object detection}, run in dedicated Docker containers orchestrated through Docker Compose. The detailed architecture, including data connections, is depicted in Fig.~\ref{fig:evaluation:use-case-architecture}.

Specifically, the test configuration encompasses a concrete OpenSCENARIO file but also general settings that define the simulation environment and sensor configuration. Thus, the two specific control layer components can interact with the CARLA server to simulate 3D lidar point cloud sensor data. These point clouds are transferred into the ROS ecosystem and subscribed by the point cloud object detection module. Resulting object detections can be visualized within RViz and stored alongside the generated raw sensor data. However, a quantitative performance assessment could be facilitated by additional components, resulting in meaningful evaluation metrics for the C-ITS module under test.

\begin{figure}[t]
    \centering
    \vspace{1.8mm}
    \includegraphics[scale=1]{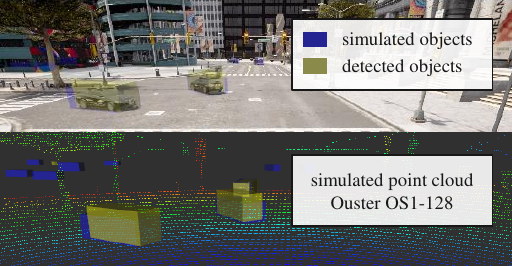}
    \caption{Visualization of simulated objects and sensor data, but also the detection results of the integrated point cloud object detection within RViz.}
    \label{fig:evaluation:use-case-impressions}
\end{figure}

Integrating such a custom, containerized module into the open simulation framework requires less effort and minimal modifications in the dedicated Docker Compose setup but directly proves general interoperability and valid interfaces with a single demo scenario. Subsequently, a visualization of the results, as in Fig.~\ref{fig:evaluation:use-case-impressions}, enables a quantitative evaluation of the tested point cloud object detection component.

\section{CONCLUSION \& OUTLOOK} 
\label{sec:conclusion-outlook}

In this paper, we have shown the benefits of an open and modular simulation framework consisting of multiple containerized building blocks which are arranged in a microservice architecture. We demonstrated the importance of simulations and scenario-based testing in all stages of the development of C-ITS software. We motivated software prototyping, data-driven development, and automated testing as essential use cases of simulations, from which we derived requirements for a generic simulation architecture. We propose usability, maintainability, interoperability, test capability, and scalability as the most essential characteristics. The result is our proposed architecture, which is implemented in the provided open-source framework CARLOS. It leverages the rich CARLA ecosystem to cover each presented use case. The framework was subsequently evaluated against the native CARLA simulator, highlighting its benefits, especially regarding scalability and maintainability.

By providing CARLOS as an open-source, easy-to-extend framework, we aim to provide utility to the ROS and CARLA communities, inviting others to contribute to this new framework. We designed CARLOS to be an open foundation for further exploration and development, especially regarding its potential for the orchestration of distributed large-scale simulations.

\addtolength{\textheight}{-16.5cm}  




\bibliographystyle{IEEEtran}
\bibliography{references/references}

\begin{thebibliography}{10}
\providecommand{\url}[1]{#1}
\csname url@rmstyle\endcsname
\providecommand{\newblock}{\relax}
\providecommand{\bibinfo}[2]{#2}
\providecommand\BIBentrySTDinterwordspacing{\spaceskip=0pt\relax}
\providecommand\BIBentryALTinterwordstretchfactor{4}
\providecommand\BIBentryALTinterwordspacing{\spaceskip=\fontdimen2\font plus
\BIBentryALTinterwordstretchfactor\fontdimen3\font minus \fontdimen4\font\relax}
\providecommand\BIBforeignlanguage[2]{{%
\expandafter\ifx\csname l@#1\endcsname\relax
\typeout{** WARNING: IEEEtran.bst: No hyphenation pattern has been}%
\typeout{** loaded for the language `#1'. Using the pattern for}%
\typeout{** the default language instead.}%
\else
\language=\csname l@#1\endcsname
\fi
#2}}

\bibitem{Groh19}
K.~Groh, S.~Wagner, T.~Kuehbeck, and A.~Knoll, ``Simulation and its contribution to evaluate highly automated driving functions,'' \emph{SAE International Journal of Advances and Current Practices in Mobility}, 2019.

\bibitem{Kloeker23}
A.~Kloeker, Y.~Liu, M.~Maier, and L.~Eckstein, ``Sim-to-real domain adaptation of infrastructure sensor lidar point clouds using generative adversarial networks,'' \emph{International Conference on Electrical, Computer and Energy Technologies}, 2023.

\bibitem{Huch23}
S.~Huch, L.~Scalerandi, E.~Rivera, and M.~Lienkamp, ``Quantifying the lidar sim-to-real domain shift: A detailed investigation using object detectors and analyzing point clouds at target-level,'' \emph{IEEE Transactions on Intelligent Vehicles}, 2023.

\bibitem{Zhong21}
Z.~Zhong, \emph{et~al.}, ``A survey on scenario-based testing for automated driving systems in high-fidelity simulation,'' \emph{Preprint}, 2021.

\bibitem{VanKempen23}
R.~van Kempen~et al., ``Autotech.agil: Architecture and technologies for orchestrating automotive agility,'' \emph{Aachen Colloquium Sustainable Mobility}, 2023.

\bibitem{Stepanyants23}
V.~G. Stepanyants and A.~Y. Romanov, ``A survey of integrated simulation environments for connected automated vehicles: Requirements, tools, and architecture,'' \emph{IEEE Intelligent Transportation Systems Magazine}, 12 2023.

\bibitem{Dosovitskiy17}
A.~Dosovitskiy, G.~Ros, F.~Codevilla, A.~Lopez, and V.~Koltun, ``Carla: An open urban driving simulator,'' \emph{1st Annual Conference on Robot Learning}, 2017.

\bibitem{Amazon24}
\BIBentryALTinterwordspacing
{Amazon.com Inc.} What’s the difference between soa and microservices? [Online]. Available: \url{https://aws.amazon.com/compare/the-difference-between-soa-microservices/}
\BIBentrySTDinterwordspacing

\bibitem{Jamshidi18}
P.~Jamshidi, C.~Pahl, N.~C. Mendonca, J.~Lewis, and S.~Tilkov, ``Microservices: The journey so far and challenges ahead,'' \emph{IEEE Software}, 2018.

\bibitem{Tasci18}
T.~Tasci, J.~Melcher, and A.~Verl, ``A container-based architecture for real-time control applications,'' \emph{International Conference on Engineering, Technology and Innovation}, 2018.

\bibitem{Baskarada20}
S.~Baškarada, V.~Nguyen, and A.~Koronios, ``Architecting microservices: Practical opportunities and challenges,'' \emph{Journal of Computer Information Systems}, 2020.

\bibitem{Hardikar21}
S.~Hardikar, P.~Ahirwar, and S.~Rajan, ``Containerization: Cloud computing based inspiration technology for adoption through docker and kubernetes,'' \emph{Second International Conference on Electronics and Sustainable Communication Systems}, 2021.

\bibitem{Lampe23}
B.~Lampe, \emph{et~al.}, ``Robotkube: Orchestrating large-scale cooperative multi-robot systems with kubernetes and ros,'' \emph{International Conference on Intelligent Transportation Systems}, 2023.

\bibitem{Xia18}
C.~Xia, Y.~Zhang, L.~Wang, S.~Coleman, and Y.~Liu, ``Microservice-based cloud robotics system for intelligent space,'' \emph{Robotics and Autonomous Systems}, 2018.

\bibitem{White17}
R.~White and H.~Christensen, ``Ros and docker,'' \emph{Part of the Studies in Computational Intelligence Book Series}, 2017.

\bibitem{Busch23}
J.-P. Busch, L.~Reiher, and L.~Eckstein, ``Enabling the deployment of any-scale robotic applications in microservice architectures through automated containerization,'' \emph{Preprint}, 2023.

\bibitem{RedHat24}
\BIBentryALTinterwordspacing
{Red Hat}. The volkswagen group builds automated testing environment. [Online]. Available: \url{https://www.redhat.com/en/success-stories/the-volkswagen-group}
\BIBentrySTDinterwordspacing

\bibitem{Forsberg92}
K.~Forsberg and H.~Mooz, ``The relationship of systems engineering to the project cycle,'' \emph{Engineering Management Journal}, 1992.

\bibitem{Akinsola20}
J.~E.~T. Akinsola, \emph{et~al.}, ``Comparative analysis of software development life cycle models (sdlc),'' \emph{Advances in Intelligent Systems and Computing}, 2020.

\bibitem{Waseem20}
M.~Waseem, P.~Liang, G.~Marquez, and A.~D. Salle, ``Testing microservices architecture-based applications: A systematic mapping study,'' \emph{Asia-Pacific Software Engineering Conference}, 2020.

\bibitem{Wachenfeld2016}
W.~Wachenfeld and H.~Winner, ``The release of autonomous vehicles,'' \emph{Autonomous Driving: Technical, Legal and Social Aspects}, 2016.

\bibitem{Lou22}
G.~Lou, Y.~Deng, X.~Zheng, M.~Zhang, and T.~Zhang, ``Testing of autonomous driving systems: Where are we and where should we go?'' \emph{ACM Joint Meeting European Software Engineering Conference and Symposium on the Foundations of Software Engineering}, 2021.

\bibitem{Fremont20}
D.~J. Fremont, \emph{et~al.}, ``Formal scenario-based testing of autonomous vehicles: From simulation to the real world,'' \emph{International Conference on Intelligent Transportation Systems}, 2020.

\bibitem{Weber23}
H.~Weber, C.~Glasmacher, M.~Schuldes, N.~Wagener, and L.~Eckstein, ``Holistic driving scenario concept for urban traffic,'' \emph{IEEE Intelligent Vehicles Symposium}, 2023.

\bibitem{Li23}
Q.~Li, \emph{et~al.}, ``Scenarionet: Open-source platform for large-scale traffic scenario simulation and modeling,'' \emph{Advances in Neural Information Processing Systems}, 2023.

\bibitem{Ika24}
\BIBentryALTinterwordspacing
{Institute for Automotive Engineering (ika)}. scenario.center. [Online]. Available: \url{https://scenario.center/}
\BIBentrySTDinterwordspacing

\bibitem{Asam24}
\BIBentryALTinterwordspacing
ASAM. Asam open x standards. [Online]. Available: \url{https://www.asam.net/standards/}
\BIBentrySTDinterwordspacing

\bibitem{Fremont18}
D.~J. Fremont, \emph{et~al.}, ``Scenic: A language for scenario specification and scene generation,'' \emph{SIGPLAN Conference on Programming Language Design and Implementation}, 2018.

\bibitem{Steimle22}
M.~Steimle, N.~Weber, and M.~Maurer, ``Toward generating sufficiently valid test case results: A method for systematically assigning test cases to test bench configurations in a scenario-based test approach for automated vehicles,'' \emph{IEEE Access}, 2022.

\bibitem{Dona22}
R.~Dona and B.~Ciuffo, ``Virtual testing of automated driving systems. a survey on validation methods,'' \emph{IEEE Access}, 2022.

\bibitem{Macenski22}
S.~Macenski, T.~Foote, B.~Gerkey, C.~Lalancette, and W.~Woodall, ``Robot operating system 2: Design, architecture, and uses in the wild,'' \emph{Science Robotics}, 2022.

\bibitem{Dobberstein2017}
J.~Dobberstein, \emph{et~al.}, ``The eclipse working group openpass – an open source approach to safety impact assessment via simulation,'' \emph{International Technical Conference on the Enhanced Safety of Vehicles}, 2017.

\bibitem{Althoff17}
M.~Althoff, M.~Koschi, and S.~Manzinger, ``Commonroad: Composable benchmarks for motion planning on roads,'' \emph{IEEE Intelligent Vehicles Symposium (IV)}, 2017.

\bibitem{Ramakrishna22}
S.~Ramakrishna, B.~Luo, C.~B. Kuhn, G.~Karsai, and A.~Dubey, ``Anti-carla: An adversarial testing framework for autonomous vehicles in carla,'' \emph{International Conference on Intelligent Transportation Systems}, 2022, focus on adversarial tets.

\bibitem{Tuncali18}
C.~E. Tuncali, G.~Fainekos, H.~Ito, and J.~Kapinski, ``Simulation-based adversarial test generation for autonomous vehicles with machine learning components,'' \emph{IEEE Intelligent Vehicles Symposium}, 2018.

\bibitem{Saigol18}
Z.~Saigol and A.~Peters, ``Verifying automated driving systems in simulation: framework and challenges,'' \emph{ITS World Congress}, 2018.

\bibitem{Bewley19}
A.~Bewley, \emph{et~al.}, ``Learning to drive from simulation without real world labels,'' \emph{IEEE International Conference on Robotics and Automation}, 2019.

\bibitem{DLR24}
\BIBentryALTinterwordspacing
Pegasus project. [Online]. Available: \url{https://www.pegasusprojekt.de}
\BIBentrySTDinterwordspacing

\bibitem{Kirchhof19}
J.~C. Kirchhof, E.~Kusmenko, B.~Rumpe, and H.~Zhang, ``Simulation as a service for cooperative vehicles,'' \emph{International Conference on Model Driven Engineering Languages and Systems Companion}, 2019.

\end{thebibliography}

\end{document}